\documentclass{isprs} % isprs class modified 23-04-2019 (Dennis Wittich)

\usepackage{booktabs}
\usepackage{subfigure}
\usepackage{setspace}
\usepackage{geometry} % added 27-02-2014 Markus Englich
\usepackage{epstopdf}
\usepackage[labelsep=period]{caption}  % added 14-04-2016 Markus Englich - Recommendation by Sebastian Brocks
\usepackage[british]{babel} 
\usepackage[hang]{footmisc}
\usepackage{lipsum}
\usepackage{tikz}
\usetikzlibrary{backgrounds, fit, positioning}
\usepackage{adjustbox}
\usepackage{stmaryrd}
\usepackage{amsmath}
\usepackage{rotating}
\usepackage{pifont}
\usepackage{xfrac}
\usepackage{multirow}
\usepackage{float}
\usepackage[hidelinks]{hyperref}
\definecolor{myblueish}{RGB}{108, 142, 191}
\definecolor{myredish}{RGB}{184, 84, 80}

\definecolor{magenta}{HTML}{FF00FF}
\definecolor{cyan}{HTML}{00FFFF}
\newcommand{\cmark}{{\color{myblueish}\ding{51}}}%
\newcommand{\xmark}{{\color{myredish}\ding{55}}}%
\newcommand{\panstrategy}{3-PAN\xspace}

\DeclareMathOperator*{\argmin}{arg\,min}

\usepackage{mathtools}
\usepackage{amsfonts}

\usepackage[capitalise]{cleveref}

\usepackage{siunitx}
\NewDocumentCommand{\rot}{O{45} O{1em} m}{\makebox[#2][l]{\rotatebox{#1}{#3}}}%
\usepackage{makecell}

\usepackage{xspace}

\newcommand{\R}{\mathbb{R}}

\newcommand{\sset}[1]{\left\{ #1 \right\}} %smart-set
\renewcommand{\vec}[1]{\boldsymbol{\mathrm{#1}}} % vector are now bold
\newcommand{\namemodel}{EOGS++\xspace}

\newcommand{\namelearnwv}{learn wv\xspace}
\newcommand{\nameflowmatching}{optical flow\xspace}

\newcommand{\constantaffineterm}{\vec{a}}

\newcommand{\code}[1]{\texttt{#1}}
\begin{document}

\title{EOGS++: Earth Observation Gaussian Splatting with Internal Camera Refinement and Direct Panchromatic Rendering}
\date{}

% KAO: Remove extra spacing

% Anonymous submissions, authors' names should not be visible
% \author{
%  Orhan Altan\textsuperscript{1}, Ian Dowman\textsuperscript{2}, Florent Lafarge\textsuperscript{3}, Clément Mallet\textsuperscript{4}, Christian Heipke\textsuperscript{5} }
\author{Pierrick Bournez\textsuperscript{1}, Luca Savant Aira\textsuperscript{2} Thibaud Ehret\textsuperscript{3}, Gabriele Facciolo\textsuperscript{1,4}}

% KAO: Remove extra newline
% Anonymous submissions, authors' affiliations should not be visible
%\address{
%	\textsuperscript{1 }ITU, Civil Engineering Faculty, 80626 Maslak Istanbul, Turkey - (oaltan, tozg, kulur, seker)@itu.edu.tr\\
%	\textsuperscript{2 }Dept.\ of Geomatic Engineering, University College London, Gower Street, London, WC1E 6BT UK - idowman@ge.ucl.ac.uk\\
%	\textsuperscript{3 }Université Côte d’Azur, INRIA – Sophia-Antipolis, France – florent.lafarge@inria.fr\\
%	\textsuperscript{4 }Univ. Gustave Eiffel, IGN-ENSG, LaSTIG – Saint-Mandé, France – clement.mallet@ign.fr\\
%	\textsuperscript{5 }Institute of Photogrammetry and GeoInformation, Leibniz Universit\"at Hannover, Germany - heipke@ipi.uni-hannover.de\\
%}
\address{
\textsuperscript{1} Universite Paris-Saclay, CNRS, ENS Paris-Saclay, Centre Borelli, Gif-sur-Yvette, France\\
\textsuperscript{2} Politecnico di Torino, Corso Duca degli Abruzzi, Torino TO, Italia
\\
\textsuperscript{3} AMIAD, Pôle Recherche, France\\
\textsuperscript{4} Institut Universitaire de France
}
% If the corresponding author is NOT the final author, always add a % space before the subsequent comma, i.e.
% first author name\textsuperscript{a,}\thanks{Corresponding author} , % second author name \textsuperscript{b}, etc.
% thanks to Niclas Borlin 05-05-2016
% information on the corresponding author should not be used any longer and has been commented out
% C. Heipke, Jan 03,2024

% the use of the information of commissions and working groups should not be used any longer and has been commented out
% C. Heipke, Sept. 20,2022
%\commission{XX, }{YY} %This field is optional. If filled, XX and YY should be replaced by adequate numbers. See https://www2.isprs.org/commissions/
%\workinggroup{XX/YY} %This field is optional.
%\icwg{}   %This field is optional.

% KAO: Use times symbol
\abstract{

Recently, 3D Gaussian Splatting has been introduced as a compelling alternative to NeRF for Earth observation, offering competitive reconstruction quality with significantly reduced training times.
 In this work, we extend the Earth Observation Gaussian Splatting (EOGS) framework to propose \namemodel, a novel method tailored for satellite imagery that directly operates on raw  high-resolution panchromatic data %and  multispectral data 
 without requiring external preprocessing.
Furthermore, leveraging optical flow techniques we embed bundle adjustment directly within the training process, avoiding reliance on external optimization tools while improving camera pose estimation.
 We also introduce several improvements to the original implementation, including early stopping and TSDF post-processing, all contributing to sharper reconstructions and better geometric accuracy.
Experiments on the IARPA 2016 and DFC2019 datasets demonstrate that EOGS++ achieves state-of-the-art performance in terms of reconstruction quality %and efficiency,
outperforming the original EOGS method and other NeRF-based methods while maintaining the computational advantages of Gaussian Splatting. Our model demonstrates an improvement from 1.33 to 1.19 mean MAE errors on buildings compared to the original EOGS models. The code is publicly available at \url{https://gardiens.github.io/EOGS2}.

}


\maketitle

%\saythanks % added 28-02-2014 Markus Englich

\section{Introduction}\label{MANUSCRIPT}

 Since the mid-20th century, the number of satellites orbiting the Earth has grown steadily, leading to an unprecedented availability of remote sensing data. Today, high-resolution panchromatic and multispectral imagery is regularly acquired over the same areas under varying conditions, and the volume of such data is expected to continue increasing in the coming years. As a result, developing scalable methods to exploit these datasets efficiently has become a key challenge for the remote sensing community. One of the main applications of such data is 3D reconstruction through photogrammetry, which aims to recover both the geometry (e.g. the digital surface models - DSM) and the appearance of the Earth’s surface from 2D images. Traditional stereovision pipelines rely on temporally consistent acquisitions and precise calibrations, which are difficult to guarantee in practice. More recent approaches, such as implicit neural representations, have shown the ability to handle diverse imaging conditions while producing accurate reconstructions. In particular, Neural Radiance Fields (NeRF)~\cite{nerf}, and its Earth-observation extensions~\cite{zhang2023sparsesat,eonerf} have demonstrated high-quality results, but remain computationally demanding.

 \begin{figure}[ht]
  %\hspace{-0.25cm}
  \begin{tabular}{c@{\hskip 0.1cm}c}
   \includegraphics[width=0.48\linewidth]{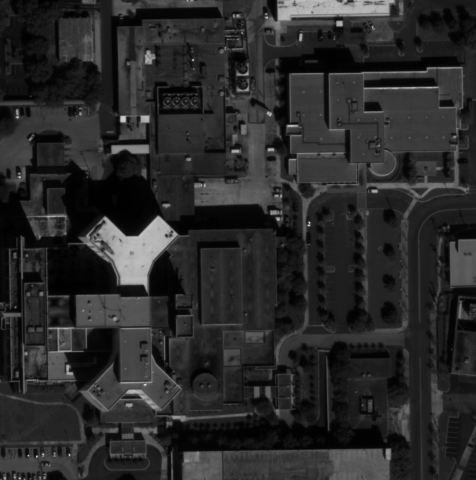} &
   \includegraphics[width=0.48\linewidth]{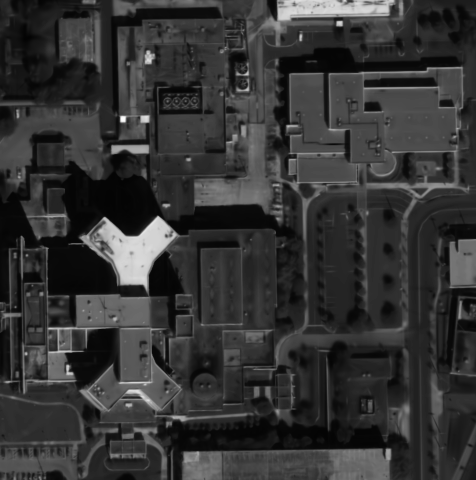} \\
   %\footnotesize{EOGS altitude} & \footnotesize{EO-NeRF altitude}\\
   Training PAN image & Rendered PAN image \\
   \includegraphics[width=0.48\linewidth]{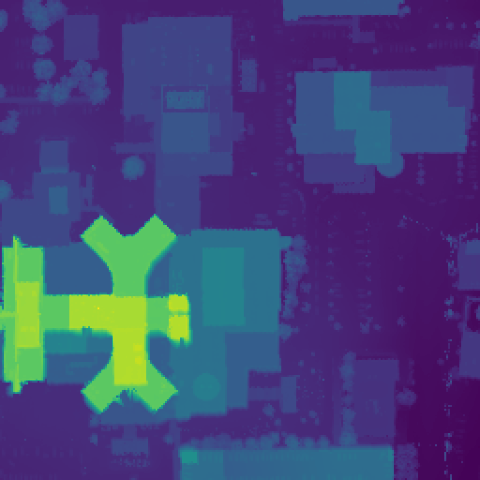} &
   \includegraphics[width=0.48\linewidth]{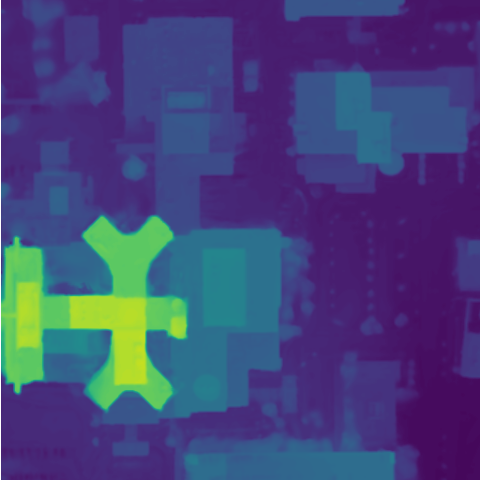} \\
    GT Lidar DSM& Rendered DSM\\
  \end{tabular}
  \vspace{-.5em}
  \caption{Qualitative result of the \namemodel pipeline.}
  \label{fig:teaser}
 \end{figure}
 
 To address these limitations, 3D Gaussian Splatting~\cite{3DGaussianSplat}, often abbreviated as 3DGS, has recently emerged as an alternative representation offering both competitive accuracy and significant computational efficiency. The EOGS~\cite{savant2025EOGS} framework successfully adapted 3DGS to satellite imagery by introducing shadow modeling, affine camera approximations, and per-camera color corrections into the pipeline. However, similar to most NeRF-based methods, it still relied on external preprocessing steps, such as pansharpening and camera poses bundle adjustment. Moreover, the original implementation was restricted to RGB data, although satellite imagery typically includes high-resolution panchromatic data alongside lower-resolution multispectral data, limiting its applicability to real-world scenarios.

%\subsection{Contribution}

The main contribution of this work is the introduction of \namemodel, a new framework for Earth observation that extends the applicability of EOGS while eliminating its dependencies. Figure~\ref{fig:teaser} illustrates qualitative results produced by our pipeline. More specifically, our contributions can be summarized as follows:
\begin{itemize}
 \item We introduce an internal bundle adjustment based on optical flow to correct camera localization errors.% \gf{This contribution is developped in Section~\ref{sec:meth-BA}.}{}

 \item We show that raw panchromatic images are sufficient to achieve a good 3D reconstruction, thus eliminating the need to pansharpen the images beforehand.%, in Section~\ref{sec:method-PANMSI},
 
 \item We improve the EOGS framework by introducing an early stopping mechanism~\cite{chung2024depth} and a truncated signed distance function-based postprocessing operation \cite{curless1996volumetric}.% \gf{We describe this part in Section~\ref{sec:optimize-eogs}.}{}
  
\end{itemize}

\begin{figure*}[ht]
	\centering
	% \begin{tikzpicture}	\node[anchor=south west, inner sep=0] (image) at (0,0) {\includegraphics[width=1\textwidth]{figures/paper/first_pic/schema/image.png}};
	% \end{tikzpicture}
  \includegraphics[width=0.9\textwidth]{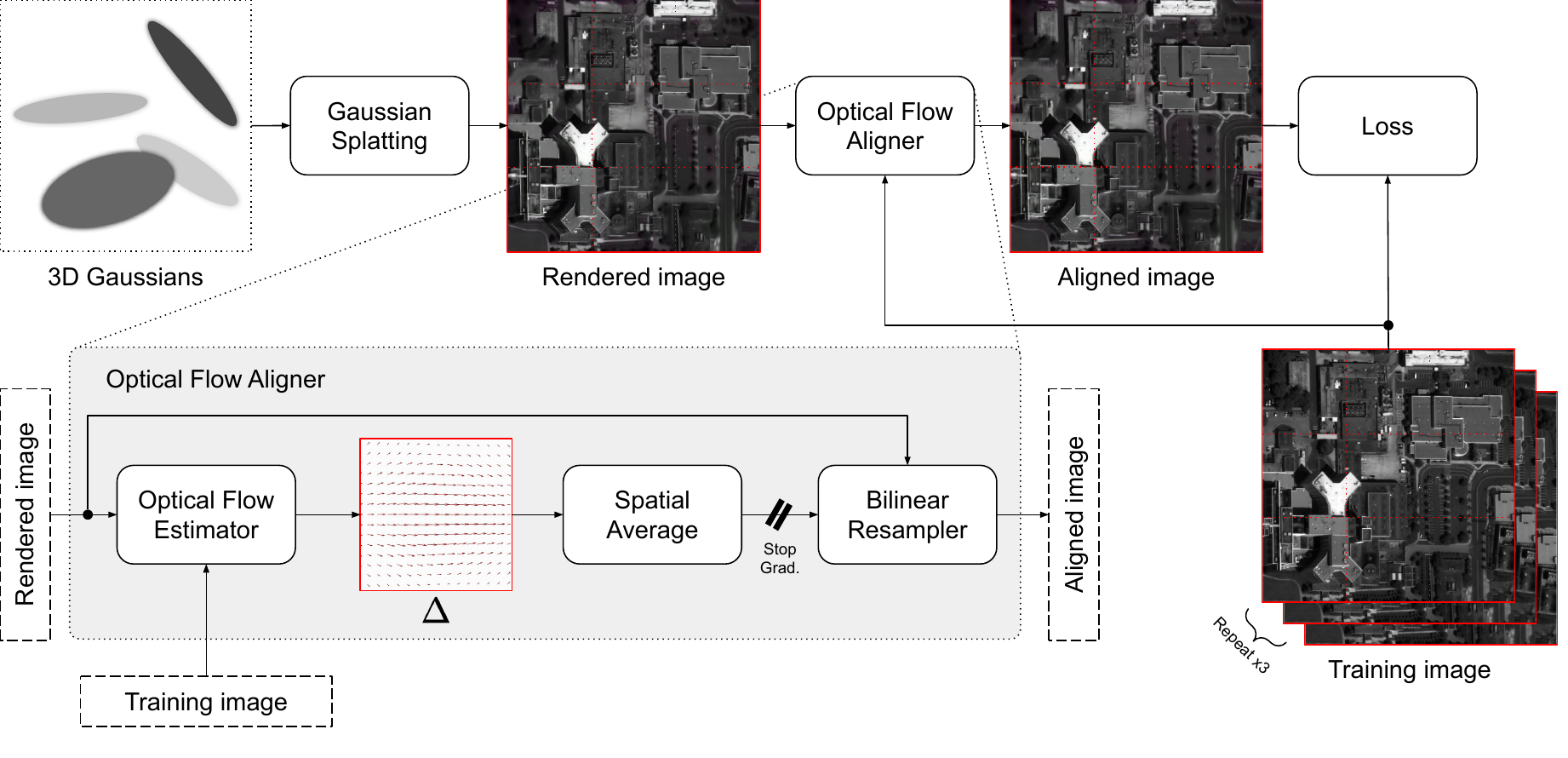}
  % \vspace{-0.4cm}
	\caption{Schematic overview of the proposed training pipeline. From the 3D Gaussian primitives, an image is rendered. The rendered image is then aligned with the training observation using an optical flow algorithm, after which the model is trained accordingly.}
	\label{fig:simu_wf} 
\end{figure*}

\section{Related Works}\label{sec:related-work}
{
  \subsection{3D reconstruction from satellite images}
  {
    Traditionally, stereoscopic 3D reconstruction from satellite imagery has relied on conventional processing pipelines, such as S2P~\cite{s2p} or Catena~\cite{krauss2013fully}%MicMAC~\cite{Micmac}
   . More recently, there has been a growing trend towards integrating deep learning models \cite{Chang_2018_CVPR} to directly regress the digital surface models. %, with approaches such as PSM or HSM. %? what is this approach? 
    While promising, these methods face limitations in terms of computational efficiency and require image acquisitions to be temporally close.
    % %The large spatial extent of acquisition areas (often spanning several square kilometers) has motivated the development of grid-based partitioning strategies to facilitate reconstruction~\cite{billouard_tile_2025
  }

  \subsection{Implicit representations for 3D reconstruction}
  {
    A different line of research focuses on implicit neural representations, such as NeRF~\cite{nerf} and Gaussian Splatting~\cite{3DGaussianSplat}, which reconstruct 3D scenes from calibrated image collections. Unlike traditional pipelines, these approaches optimize a continuous scene representation and have demonstrated improved performance across diverse imaging conditions~\cite{yan2024gs}. 

    Adapting implicit methods to satellite imagery, however, presents additional challenges due to sparse viewpoints, limited variation in viewing angles across acquisitions, and varying illumination. Several works have addressed these limitations by extending NeRF-like models to pushbroom cameras \cite{derksen2021shadow,mari2022sat}, explicitly modeling shadows \cite{eonerf,behari_sundial_2024}, or filtering transient elements to improve reconstruction robustness \cite{huang_skysplat_2025}. 

    While NeRF-based techniques have been actively explored, literature on Gaussian Splatting in the satellite domain remains limited~\cite{savant2025EOGS,huang_skysplat_2025}. This approach achieves performance comparable to NeRF-like methods while requiring shorter training times.
  }

  \subsection{ Pansharpening}
  {
    Pansharpening aims to fuse high-resolution panchromatic images with low-resolution multispectral images to produce high-resolution multispectral outputs.

    Classical approaches can be grouped into three main categories: Component Substitution (CS), Multi-Resolution Analysis (MRA), and variational methods. CS methods rely on spectral transformations such as IHS~\cite{carper1990use} or PCA~\cite{ghadjati2019novel}, while MRA techniques use spatial filters (e.g., Laplacian pyramids~\cite{li1995multisensor}) to extract high-frequency details from the panchromatic image and inject them into the multispectral image. Variational approaches~\cite{palsson2013new,meng2020large} formulate the fusion as an energy minimization problem to ensure spectral and spatial consistency.

    More recently, deep learning-based methods such as PNN~\cite{PNN} have gained attention. However, these models often face generalization issues across different sensors or geographic regions.

    In the context of inverse rendering, most pipelines~\cite{mari2022sat} perform pansharpening as a preprocessing step prior to 3D reconstruction. Alternative strategies, however, directly leverage multispectral data, either by predicting spectral channels independently~\cite{Nerfpansharpalt} or by explicitly modeling inter-band dependencies~\cite{Eonerfpansharp}. 
  }

  \subsection{ Bundle adjustment}
  {
    Accurate 3D reconstruction requires reliable camera poses. While satellite providers supply approximate orientations, these are often affected by geolocation errors~\cite{grodecki2003block}, which can degrade reconstruction quality~\cite{eonerf}. Bundle Adjustment (BA)~\cite{triggs1999bundle} is commonly used to refine these estimates by minimizing reprojection errors~\cite{mari2021generic}. 

    Classical BA pipelines rely on interest-point matching (e.g., SIFT~\cite{lowe1999object}), but are computationally expensive and sensitive to outliers~\cite{szeliski2022computer}. More recent approaches embed pose refinement within learning-based frameworks, jointly optimizing poses and geometry, for instance in NeRF~\cite{barf,eonerf}. In the context of Gaussian Splatting, BA has been explored mainly in SLAM~\cite{yan2024gs} and for correcting motion blur~\cite{deng2025ebad}, with recent work proposing BA directly on Gaussian parameters~\cite{zhang2024gbr}.
  }
}

\section{Gaussian splatting for earth observation}\label{sec:pres-eogs}
{

 {Our approach is based on Earth Observation Gaussian Splatting (EOGS)~\cite{savant2025EOGS}, which is an extension of 3D Gaussian Splatting (3DGS)~\cite{3DGaussianSplat} to satellite images.
 In this section, we present in detail the main ideas of these two methods.
 The core idea behind 3DGS is that, given a set of $N$ images and their corresponding camera models, a set of $K$ Gaussian 3D primitives is optimized to recover the appearance of the scene.
 In practice, this collection of primitives $\sset{\gamma_k}_{k=1}^{K}$ is parametrized by their respective position $\vec{\mu}_k \in \R^3$, covariance $\Sigma_k \in \mathcal{M}^{3 \times 3}(\R)$, color features $\vec{c}_k \in \R^3$ and opacity $\alpha_k \in [0,1]$. Each primitive has an associated Gaussian density function defined as
 \begin{equation}
  \mathcal{G}_k (\vec{x}) = \exp \left\{ -\tfrac{1}{2} (\vec{x}-\vec{\mu_k})^T \Sigma^{-1}_k (\vec{x}-\vec{\mu_k})\right\}.
 \end{equation}
 
 The link with the observations and the set of primitives is given by the rendering pipeline. All primitives are first projected onto the image plane thanks to the camera model. The projected Gaussians $\mathcal{G}^{\mathcal{C}}_k$ are then sorted front-to-back (based on their center position) and aggregated using an alpha-compositing scheme:
\begin{equation}\label{eq:rendering}
  I^{\mathcal{C}}(\vec{u}) = \sum_{k=1}^K \vec{c}_k \omega^{\mathcal{C}}_k (\vec{u}),
 \end{equation}
 where $\vec{u}$ is the coordinate of a pixel on the image plane $I^{\mathcal{C}}$ associated to the camera $\mathcal{C}$ and $\omega^{\mathcal{C}}_k$ is the accumulated opacity of the $k$-th Gaussian at pixel $\vec{u}$, defined as
 \begin{equation}
  \omega^{\mathcal{C}}_k (\vec{u}) = \alpha_k \mathcal{G}^{\mathcal{C}}_k (\vec{u}) \prod_{j=1}^{k-1} (1 - \alpha_j \mathcal{G}^{\mathcal{C}}_j (\vec{u})).
 \end{equation}

  The learning problem is therefore to find the set of $K$ Gaussian primitives that best approximates the $N$ images, with the rendering process of \cref{eq:rendering}. This can be formulated as:
\begin{equation}\label{eq:learningproblem}
 \argmin_{\sset{\gamma_k}_{k=1}^{K}} \sum_{n=1}^N \ell_1(I^{\mathcal{C}_n} - I^n) + \lambda \ell_{D-SSIM}(I^{\mathcal{C}_n}, I^n),
 \end{equation}
 where $\lambda$ is the weight term, $I^n$ is the $n$-th input observation, corresponding to camera $\mathcal{C}_n$, $I^{\mathcal{C}_n}$ is the corresponding synthesized view,  $\ell_1$ is the pixelwise
$L_1$ reconstruction loss, and $\ell_{D\text{-}SSIM}$ is the differentiable SSIM-based
loss introduced in the original 3D Gaussian Splatting framework~\cite{3DGaussianSplat}.

% 1. Camera model
 EOGS extends the original 3DGS framework to the satellite imaging setting. We summarize here the key differences with the original 3DGS framework that will be useful for the rest of paper. First of all, the camera model used is different. Indeed, 3DGS is designed to work with pinhole camera model, which is the standard model for most consumer cameras. However this camera model is not appropriate in the case of satellite images where the camera model is usually represented with a Rational Polynomial Coefficient (RPC) camera model~\cite{de2014stereo}. Because EOGS argues that the projection step is computationally inefficient when applied to RPCs, it instead employs an approximate affine camera model $\mathcal{A}$.
 Hence, each camera is described as an affine function $\mathcal{A} : \R^3 \to \R^2$, mapping 3D points $\vec{x}$ to 2D image coordinates $\vec{u} = \mathcal{A}(\vec{x}) = A \vec{x} + \vec{a}$ with $A \in \R^{2 \times 3}$ and $\constantaffineterm \in \R^2$. In that case, The projection operation assigns to each Gaussian a 2D Gaussian density function, with mean $\vec{\mu}^{\mathcal{A}}_k = \mathcal{A}(\vec{\mu}_k) \in \R^2$ and covariance $\Sigma^{\mathcal{A}}_k = A \Sigma_k A^T \in \mathcal{M}^{2 \times 2}(\R)$.\\ 
%3. ELEVATIOn
Similarly to mesh-based variants of 3DGS, e.g. ~\cite{guavsdon2025milo}, EOGS aims to recover the 3D geometry of the scene. To this end, it defines an elevation rendering that uses the true elevation rather than RGB values: \begin{equation} E^{\mathcal{A}} (\vec{u}) = \sum_{k=1}^{K} \mathcal{E}(\mu_k)\, \omega^{\mathcal{A}}_k(\vec{u}), \end{equation} where $\mathcal{E} : \mathbb{R}^3 \to \mathbb{R}$ is an affine mapping that returns the real altitude, expressed in meters. In the original implementation, this elevation rendering is used to evaluate the quality of the reconstructed 3D geometry.

 % 2. Color correction + shadow model
 EOGS also improves the alpha-compositing scheme that was described in~\eqref{eq:rendering}. It introduces the concept of shadow mapping inside the rendering pipeline. Modeled as a binary mask $l^{\mathcal{A}}$, it represents the visible regions where lighting coming from the sun is occluded by the scene. % Given that we only need shadows for the rendering function, I don't think that it's worth getting into more details but it can be easily be added if necessary.
 It also introduces a per-camera color-correction, as satellite image colors can change due to difference in atmospheric conditions at the acquisition dates. To model this effect, the authors use a learnable $d \times 3$ per-camera affine transformation, $\vec{\phi}^{\mathcal{A}}$, that maps Gaussian attributes to color channels appropriate for the given acquisition.
  This leads to the following alpha-compositing scheme
 \begin{equation}\label{eq:renderingEOGS}
   I^{\mathcal{A}}(\vec{u}) = l^{\mathcal{A}}(\vec{u}) \sum_{k=1}^K \vec{\phi}^{\mathcal{A}}(\vec{f}_k) \omega^{\mathcal{A}}_k (\vec{u}).
  \end{equation}
 % 3. Regularization + view-consistency
 Finally, the authors also add a regularization term to encourage view consistency. Given a small perturbation of a camera model, view consistency is encouraged by penalizing differences between the perturbated view and the original view (after alignment and masking of occluded regions). Two terms are proposed, one for the elevation ($\ell_{ec}$) and one for the color ($\ell_{cc}$). Additional terms to promote sparsity and opaqueness (referred as $R$ in the following) are also included~\cite{savant2025EOGS}. The complete EOGS learning problem can then be formulated as
\begin{align}\label{eq:learningproblemEOGS}
 & \argmin_{\sset{\gamma_k}_{k=1}^{K},\sset{\phi^{\mathcal{A}_n}}_{n=1}^{N}} \sum_{n=1}^N \ell_1(I^{\mathcal{A}_n} - I^n) + \lambda_1 \ell_{D-SSIM}(I^{\mathcal{A}_n}, I^n) \nonumber\\
 & + \lambda_2 \ell_{ec}(I^{\mathcal{A}_n}) + \lambda_3 \ell_{cc}(I^{\mathcal{A}_n}) + \lambda_4 R(\sset{\gamma_k}_{k=1}^{K}, \mathcal{A}_n,l^{\mathcal{A}_n}),
 \end{align}
 with $\lambda$s the different weighting terms for each corresponding loss. In the following, we define the \textit{photometric loss} $\ell$ as the combination of the $\ell_1$ loss and the $\ell_{D-SIMM}$ loss.
 
}

\section{Method}\label{Method}
{
 The proposed \namemodel extends and improves EOGS during preprocessing, training and postprocessing phases. 
 The required preprocessing steps in EOGS, such as bundle adjustment and pansharpening, are integrated in \namemodel, as explained respectively in Sections~\ref{sec:meth-BA} and~\ref{sec:method-PANMSI}. This enables \namemodel to have virtually zero required preprocessing steps.
 The training phase of EOGS is also improved with the introduction of opacity reset and early stopping, as explained in Section~\ref{sec:optimize-eogs}.
 Finally, a postprocessing step is proposed to improve the final DSM prediction, as explained in Section~\ref{sec:meth-refine-mae-comp}. The training pipeline is illustrated in Figure~\ref{fig:simu_wf}.

 \subsection{Bundle adjustment}\label{sec:meth-BA}
 {
  Bundle adjustment (BA) in satellite imagery is commonly used to refine camera models estimates by minimizing reprojection errors~\cite{grodecki2003block}, and it is an essential preprocessing step in EOGS to ensure good reconstruction results, as demonstrated in Table~\ref{tab:RPC_quantitative}.
  \namemodel integrates bundle adjustment directly within the training process, eliminating the need for external BA software. This is achieved by leveraging optical flow to align rendered and reference images.

  First, the image $I^{\mathcal{A}}$ is rendered, using \cref{eq:rendering}. Then an off-the-shelf optical flow estimation model computes a displacement field $\Delta: \R^2 \to \R^2$, between $I^{\mathcal{A}}$ and the corresponding training image $I$, such that the pixel $\vec{u}+\Delta(\vec{u})$ in $I^{\mathcal{A}}$ corresponds to the pixel $\vec{u}$ in $I$.
  
  Since camera miscalibration primarily manifests as pixel offsets in the context of satellite imagery~\cite{mari2021generic}, a constant averaged version of $\Delta$ is considered, namely $\Bar{\Delta} \in \R^2$. $\Bar{\Delta}$ is used to bilinearly resample the rendered image: $\hat{I}^{\mathcal{A}} \approx I^{\mathcal{A}}(\vec{u}+\Bar{\Delta})$. This is equivalent to consider the shifted version of the rendered image that better matches the training image. 
  To keep the optical‑flow estimator isolated from the image optimizer, we block its gradients by applying a stop‑gradient to $\Bar{\Delta}$ so that no back‑propagation flows through it.
%  Furthermore, to prevent gradients from the optical flow estimator flowing into the image optimization,  we don't compute gradients through the optical flow estimator; therefore, a stop-gradient operation is applied to $\Bar{\Delta}$.
  Finally, the role of $I^{\mathcal{A}_n}$ is replaced by $\hat{I}^{\mathcal{A}_n}$ in \cref{eq:learningproblemEOGS}.

 }

 \subsection{Handling panchromatic images}\label{sec:method-PANMSI} % Expliquer solution?
 {
  While EOGS handles only RGB images, satellite imagery typically includes high-resolution panchromatic (PAN) images alongside lower-resolution multispectral (MSI) images. Hence a preprocessing step, named pansharpening, is required to transform the MSI images into high-resolution RGB images before inputting them into the model. This step, however, can introduce artifacts and inconsistencies, especially when the PAN and MSI images are not perfectly aligned or captured under different conditions.
   \namemodel discards the low-resolution MSI images and reconstructs the scene only from the original high-resolution PAN data. The single-channel panchromatic training images are repeated into a three-channel RGB representation, and the rendered outputs $I^{\mathcal{A}}$  are compared directly to the replicated training images. This strategy is referred to as \panstrategy.
  We acknowledge that this strategy disregards the spectral information from the MSI images, which could be valuable for certain applications. However, as we will show experimentally in Section~\ref{sec:res-pansharpen}, the inclusion of MSI data does not yield a significant gain for surface model estimation.% this is counterintuitively beneficial.
 }

\begin{figure}[t]
\centering
\setlength{\tabcolsep}{0.6pt}
\renewcommand{\arraystretch}{0.0}
\begin{tabular}{@{}ccc@{}}
% &
{Without reset opacity} & {With reset opacity}& \\[4pt]
% \adjustbox{angle=90, raise=0.09\linewidth, origin=c}{IARPA 001} &
\includegraphics[width=0.5\linewidth]{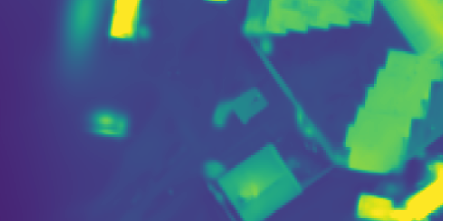} &
\includegraphics[width=0.5\linewidth]{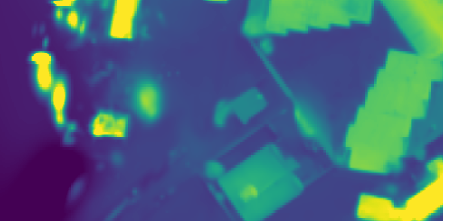} & 
\includegraphics[width=0.08\linewidth]{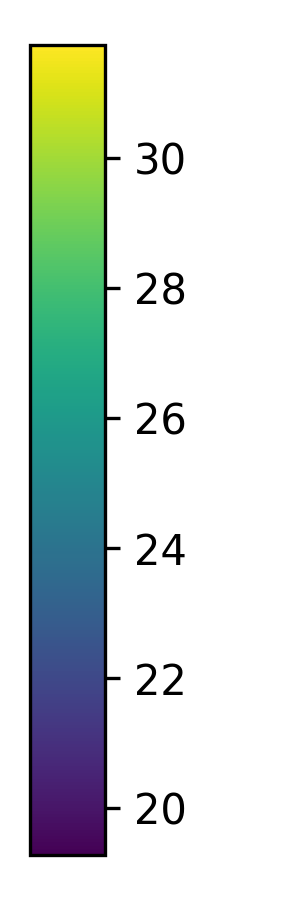} \\[1.2pt]
% \adjustbox{angle=90, raise=1.3cm, origin=c}{IARPA 002} &
\includegraphics[width=0.5\linewidth, trim=0 20 0 120, clip]{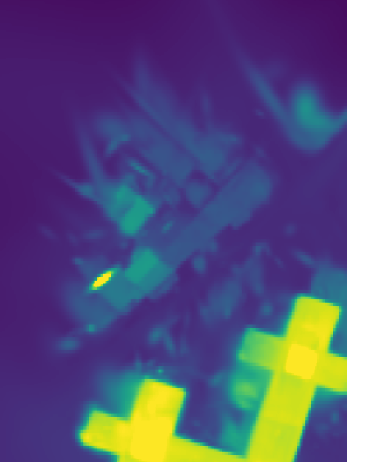} &
\includegraphics[width=0.5\linewidth, trim=0 20 0 120, clip]{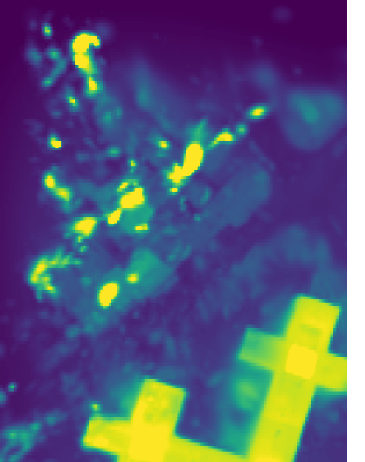} & 
\includegraphics[width=0.12\linewidth]{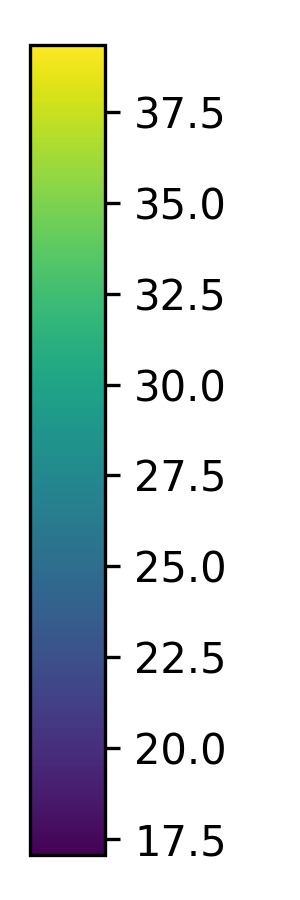} \\[1.2pt]
% \adjustbox{angle=90, raise=0.15352\linewidth, origin=c}{JAX 214} &
\includegraphics[width=0.5\linewidth]{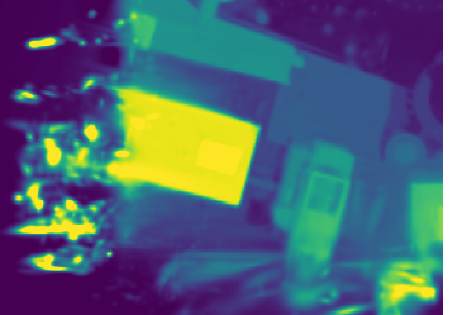} &
\includegraphics[width=0.5\linewidth]{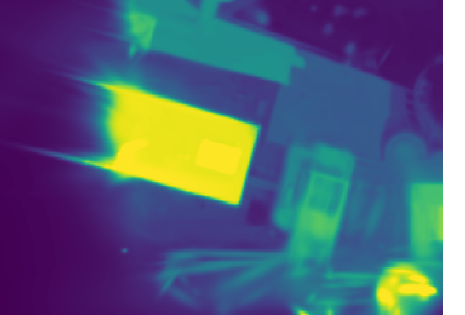}& 
\includegraphics[width=0.12\linewidth]{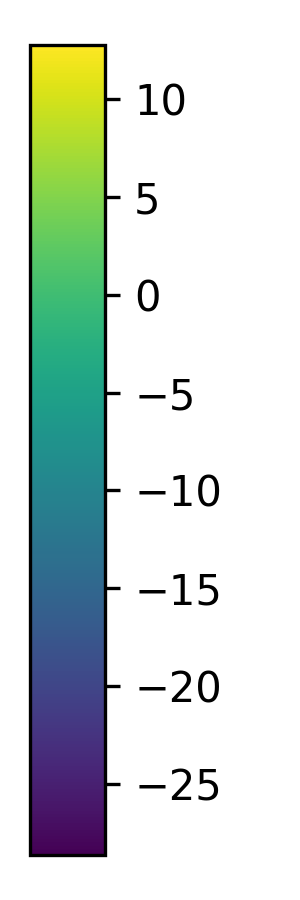} 
\end{tabular}
\vspace{-.5em}
\caption{Zoomed-in DSM comparison for respectively the IARPA 001, IARPA 002, and JAX 214 with and without opacity reset. The reset operation helps eliminate Gaussian floaters. 
\label{fig:DSM_reset_opacity}}
\end{figure}
 
 \subsection{Opacity reset and early stopping}\label{sec:optimize-eogs}
 {

  % \TODO{Sell the story of opacity reset and early stopping as a bundle.}
  %The original EOGS implementation~\cite{savant2025EOGS} was primarily designed for speed. As such, it did not fully leverage all the techniques introduced in the original 3DGS paper~\cite{3DGaussianSplat}, which could enhance reconstruction quality and stability.
  We observed that while EOGS converges rapidly, the final reconstructions sometimes lack sharpness and fine details as shown in  Figure~\ref{fig:DSM_reset_opacity}. To address this, we reintegrated a technique from the original 3DGS paper, called \textit{opacity reset}, which involves cyclically resetting the opacity of all Gaussians every \num{3000} iterations, i.e., $\alpha_k \leftarrow \min(\alpha_k,0.05)$. 
  In order to fully benefit from this mechanism, we also increase the maximum number of iterations from \num{5000} to \num{40000}. However, we noticed that by enabling the reset opacity mechanism during the optimization process, the photometric loss $\ell$ of \cref{eq:learningproblemEOGS} exhibits a strong non-monotonic behavior: it first decreases, but after a number of iterations starts to increase. We speculate that this is caused by the regularizers in \cref{eq:learningproblemEOGS} which take hold of the training dynamics. To mitigate this issue, we implemented an early stopping mechanism based on photometric loss $\ell$ itself. Training is therefore halted as soon as the photometric loss $\ell$ starts to increase, as a minimum in the photometric loss indicates a high-fidelity reconstruction. Thus, this approach enhances sharpness while minimizing the loss of fine and useful details. In practice, this criterion leads to convergence after approximately \num{10\,000} to \num{20\,000} iterations
 
 }

% imposing consistency across viewa (merge depth predictions) %Refining the DSM reconstruction 
  \subsection{Imposing explicit depth consistency across views}\label{sec:meth-refine-mae-comp}
 {
  % In the original EOGS implementation, a virtual nadir-view camera was used to compute elevation maps for MAE evaluation. However, since this camera is not part of the training process, it can introduce artifacts, particularly when training views deviate significantly from the nadir configuration.

  % To address this, we refine the MAE computation. A DSM is estimated from each training view and these are combined to fit a signed distance function (SDF), providing a more consistent and representative surface for evaluation. \TODO{Complete this section with implementation detail}

  In the original EOGS implementation, a virtual nadir-view camera was used to compute the final elevation maps for MAE evaluation. However, since this camera is not part of the training process, it can introduce artifacts, particularly when training views deviate significantly from the nadir configuration. %\gf{}{[TODO: refer the EOGS implicit regularizer explaining that it does not necessarily achieves cross-view consistency.]}

  To address this, we propose an optional post-processing step based on the truncated signed distance function (TSDF) fusion technique, akin to many current methods {\cite{huang20242d}. Our method aggregates depth maps from all training views to create a more accurate and consistent digital surface model. The process involves rendering depth maps from each training view using the final optimized Gaussian primitives, then fusing these depth maps into TSDF, with a fixed truncation margin and with weights proportional to the angle between the training view direction and an estimated surface normal, as described in \cite{curless1996volumetric}. Finally, we extract a fixed-resolution voxelization of the TSDF using marching cubes \cite{lorensen1998marching}, resulting in a mesh. We enhance this mesh by removing isolated voxels (stemming from artifacts known as floaters in the Gaussian Splatting literature) and bottom-up filling holes. The final DSM is then obtained by sampling the height of this refined mesh at each ground position.
  
 %  We remark that, while this post-processing step improves the MAE evaluation, it does also create a mesh that could be used for other applications beyond mere evaluation.
 %  \todo{Pierrick : Shouldn't we add some equations? Or it is too heavy?
 %  Maybe the last sentence could be removed or moved to conclusion.Because reviewers could ask why we didn't do it }
 % }

}

% \begin{figure}[t]
%   \centering
%   \includegraphics[width=0.38\linewidth]{figures/paper/res-ba/final_rpcraw_image.png}
%   \includegraphics[width=0.38\linewidth]{figures/paper/res-ba/final_learnwv_image.png}\\
%   \includegraphics[width=0.38\linewidth]{figures/paper/res-ba/final_flowmatch_image.png}
%   \includegraphics[width=0.38\linewidth]{figures/paper/res-ba/final_rpcba_image.png}
%   \caption{From top-left to bottom-right: rendered images on IARPA\_001 using the raw RPC, the \namelearnwv method, the optical flow correction, and the bundle-adjusted cameras on the panchromatic images. Notice how the images become sharper}
%   \label{fig:RPC_qualitative_comparison_rendering}
% \end{figure}

\section{Experiments}
{

 \subsection{Datasets}

We evaluate \namemodel under experimental conditions similar to those used in the original EOGS article. The datasets employed are extracted from the 2016 IARPA Multi-View Stereo 3D Mapping Challenge~\cite{le20192019}, denoted as IARPA2016, and the 2019 IEEE GRSS DATA Fusion Contest~\cite{US3DDataset}, denoted as DFC2019. Overall, these datasets consist of 7 areas of interest and include multi-date non-orthorectified PAN and MSI WorldView-3 satellite observations. Each area covers a terrain of $256 \times 256$ m$^2$, with a spatial resolution ranging from 30 to 50~cm per pixel for the PAN images and four times coarser for the MSI images. Each site is observed from approximately 10 to 30 different points of view.

\subsection{Implementation details }
 The implementation builds upon the publicly available EOGS code. 
\namemodel uses \code{RAFT\_small}~\cite{raft} as the optical flow estimator for camera calibration, that represents a good tradeoff between speed and accuracy. %\luca{}{We remark that this approach does not require to compute the gradient of the optical flow estimator with respect to the loss, so a stop-grad operation is applied on $\Bar{\Delta}$.}

Similarly to EOGS, we report the Mean Absolute Error (MAE) between the LiDAR scan provided in the dataset and the predicted elevation map obtained either as a nadir depth rendering or through a TSDF post-processing step, as described in Section~\ref{sec:meth-refine-mae-comp}. Because foliage conditions vary across seasons and over time, and the reconstruction primarily targets buildings, the MAE is computed using a foliage mask, following the practice introduced by EOGS.

\subsection{Internal bundle adjustment}\label{sec:res-internal-ba}

Table~\ref{tab:RPC_quantitative} shows multiple ways %\gf{to do RPC bundle adjustment}
{of handling errors in the camera pointing}, by comparing raw RPCs (\textit{none}), internally learned RPC corrections (\textit{learn wv}), the proposed optical-flow-based method (\textit{optical flow}) and externally bundle adjusted RPCs (\textit{external B.A.}).
During the early development phase of \namemodel, we implemented a straightforward bundle adjustment algorithm, referred to as \namelearnwv, which involves applying backpropagation to the affine projection matrix.
In this strategy, both the Gaussian positions and the projection matrices of each camera are optimized during training. This enables the model to implicitly recalibrate the camera models. Similarly to the proposed optical flow approach, we restrict the optimization to the constant terms $\constantaffineterm$ of the affine transformation, in order to learn just a per-camera constant shift in the image plane. The original implementation did not support gradient descent with respect to projection matrices. We extended the framework to allow gradient-based updates for affine camera models.
% \gf{}{[Q: this ends like this? what happens with the gradient-based updates??]}

\begin{table}[!ht] 
\centering
\setlength{\tabcolsep}{3pt}
\renewcommand{\arraystretch}{1.1}
\begin{tabular}{rccc}
B.A. algorithm              & JAX $\downarrow$  & IARPA $\downarrow$    & Mean $\downarrow$ \\ \hline
None                        & 1.33              & 2.54                  & 1.93              \\
Learn wv                    & 1.21              & 2.43                  & 1.82              \\
Optical flow                & 1.23              & 1.50                  & 1.36              \\ 
External B.A. (Reference)   & \textbf{1.19}     & \textbf{1.46}         & \textbf{1.33}     \\ \hline
\end{tabular}
\caption{
Raw RPC handling comparison. Reported metric is the mean absolute error of the predicted elevation map [meters], when ignoring foliage areas. To ensure a fair comparison, all methods use the same pre-computed pansharpened images and are trained for \num{5000} iterations
} 
\label{tab:RPC_quantitative}
\end{table}

\begin{figure}[!hb]
  \centering
  \setlength{\tabcolsep}{1pt}
  \renewcommand{\arraystretch}{0.0}
  \begin{tabular}{@{}lcc@{}}
  & IARPA 002 & IARPA 003 \\[4pt]

  \adjustbox{angle=90, raise=0.2\linewidth, origin=c}{None}
  &\includegraphics[width=0.43\linewidth]{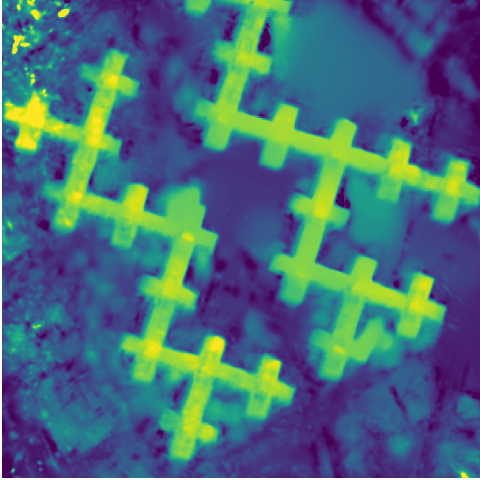}
  &\includegraphics[width=0.43\linewidth]{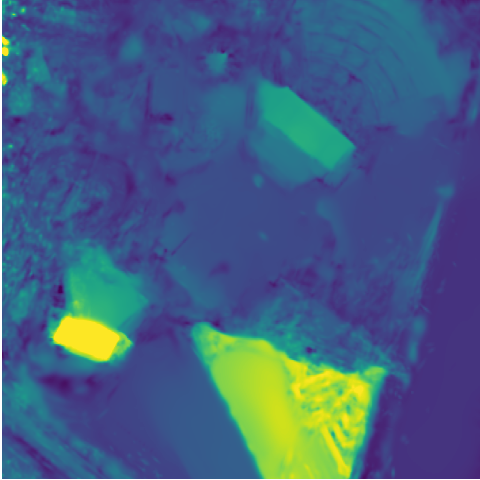} \\
  
  \adjustbox{angle=90, raise=0.2\linewidth, origin=c}{Learn wv}

  &\includegraphics[width=0.43\linewidth]{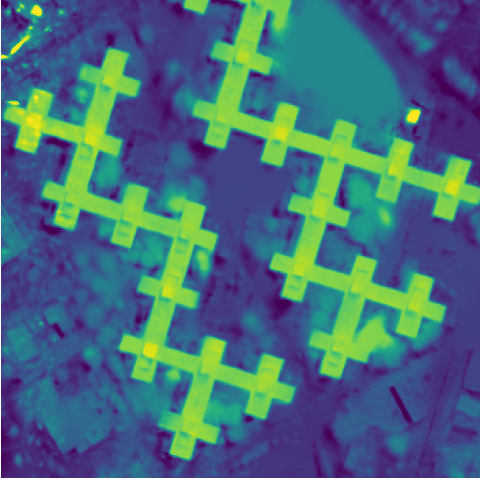}
  &\includegraphics[width=0.43\linewidth]{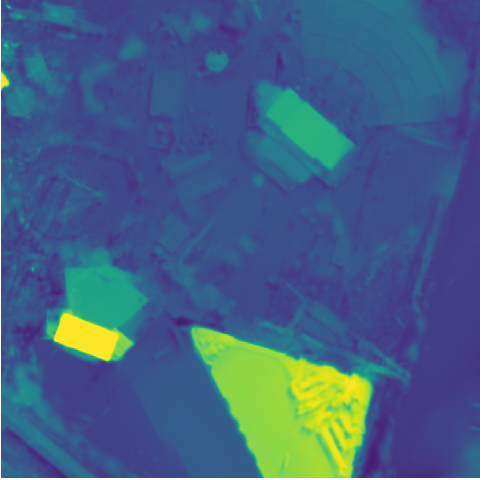} \\

  \adjustbox{angle=90, raise=0.2\linewidth, origin=c}{Optical flow}
  &\includegraphics[width=0.43\linewidth]{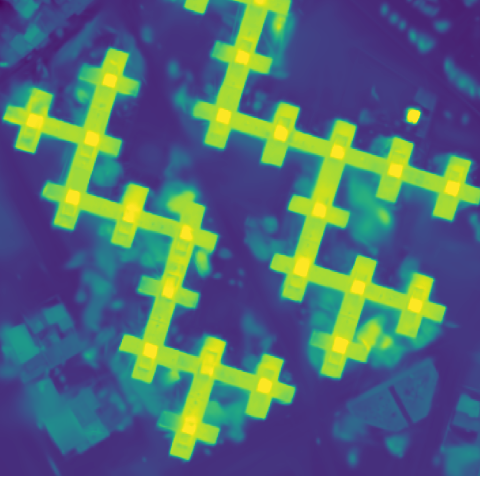}
  &\includegraphics[width=0.43\linewidth]{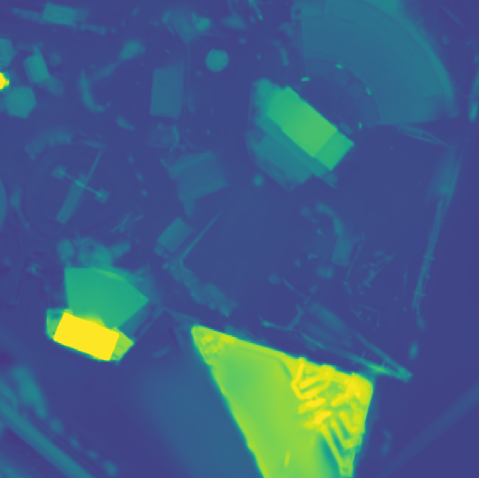} \\

  \adjustbox{angle=90, raise=0.2\linewidth, origin=c}{External B.A.}
  &\includegraphics[width=0.43\linewidth]{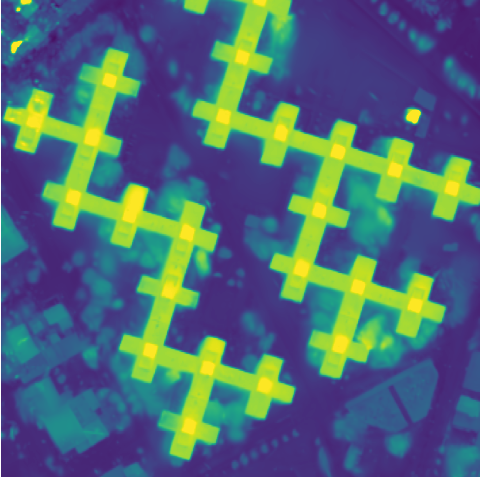}
  &\includegraphics[width=0.43\linewidth]{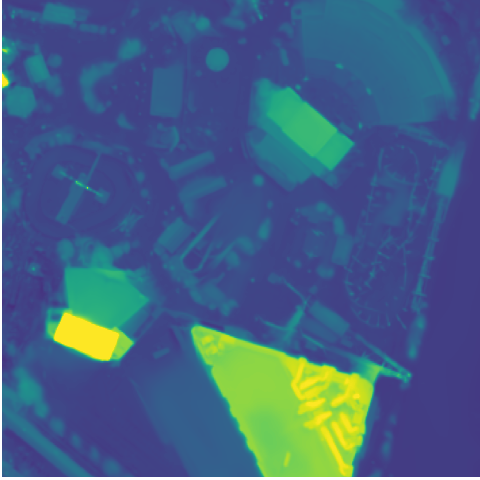} \\

  \adjustbox{angle=90, raise=0.2\linewidth, origin=c}{GT Lidar DSM}
  &\includegraphics[width=0.43\linewidth]{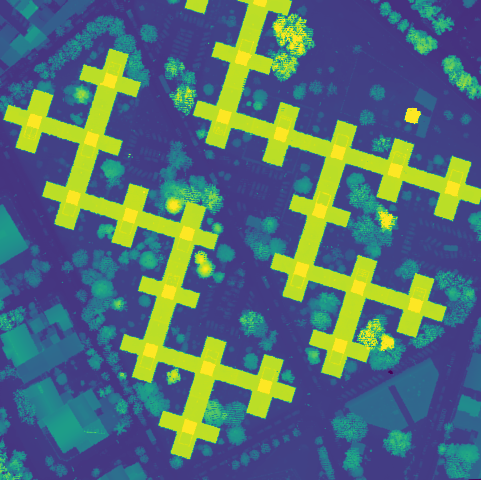}
  &\includegraphics[width=0.43\linewidth]{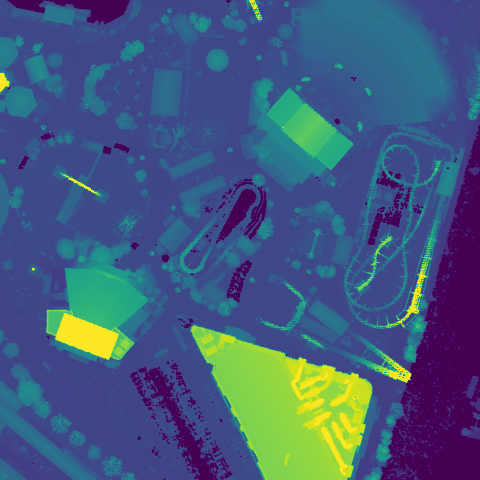}
  \\
  & \includegraphics[width=0.43\linewidth]{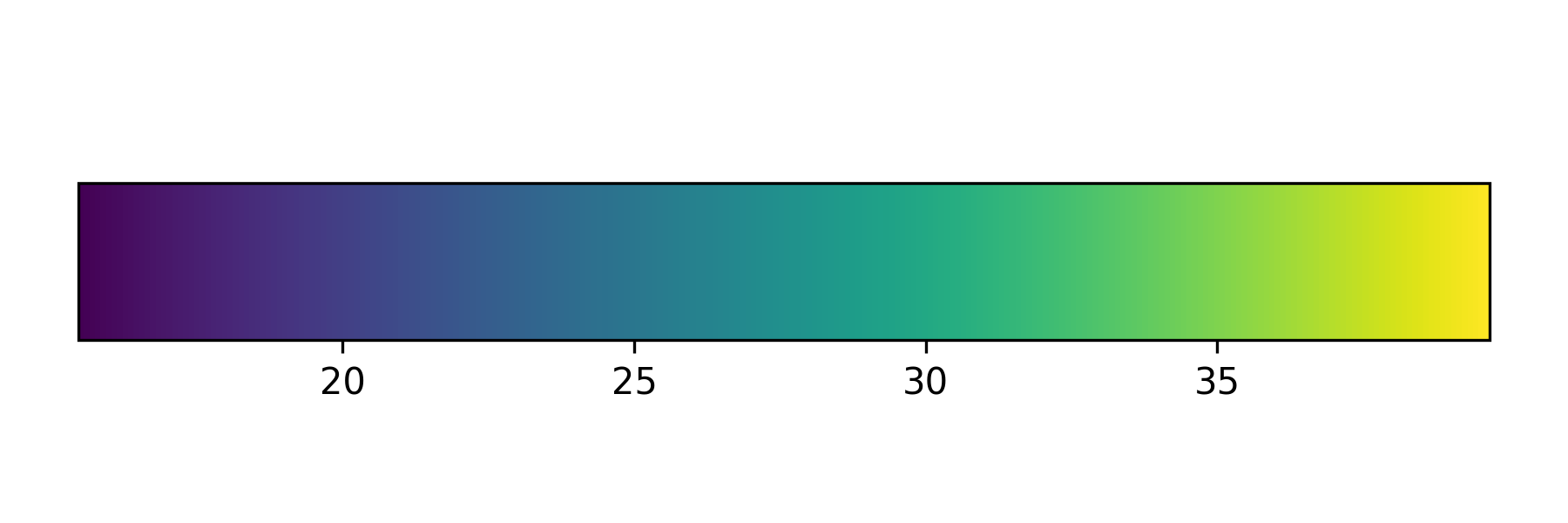}& 
  \includegraphics[width=0.43\linewidth]{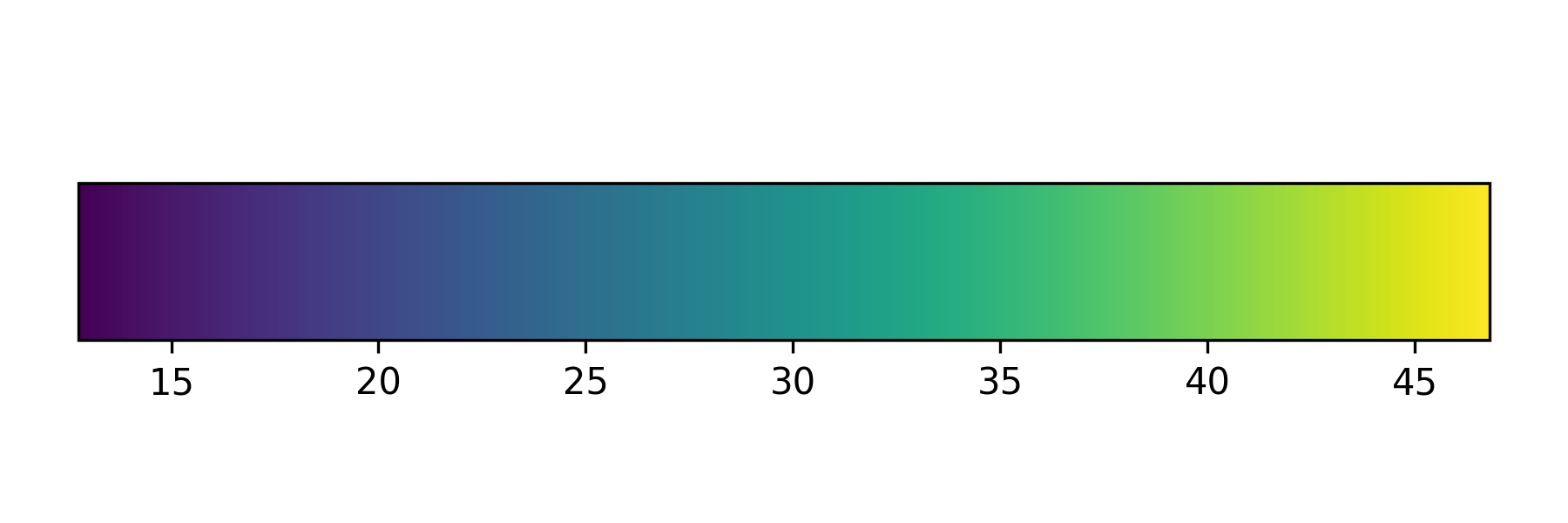}
  
  \end{tabular}

    \caption{Qualitative comparison of different methods for handling errors in the camera pointing (\textit{cf.} \cref{tab:RPC_quantitative}) on IARPA~002 and IARPA~003 scenes.}
  \label{fig:visual-DSM-ba}
\end{figure}

 While the \namelearnwv method is able to correctly align and render images (see Figure~\ref{fig:RPC_qualitative_comparison_rendering}), its MAE performance does not show significant improvement compared to the raw RPC initialization. We hypothesize that the joint optimization of Gaussian parameters and affine matrices leads to convergence toward bad local minima.
Conversely, the \nameflowmatching algorithm effectively registers the images, achieving comparable results (within 3 cm) when compared to bundle-adjusted cameras (see the DSM in Figure~\ref{fig:visual-DSM-ba}). Eventually the best performance is obtained with externally bundle-adjusted cameras, which we attribute to the improved stability of explicit triangulation in the bundle adjustment procedure compared with RGB-based optimization.

\subsection{The need for multispectral images}\label{sec:res-pansharpen}
%  {
%   Table~\ref{tab:PAN_quantitiative} \luca{summarizes the results. We also explored two alternative approaches that appeared natural for improving the reconstruction, but found that using only the panchromatic (PAN) images yielded superior results. The first approach, referred to as \textit{Brovey}, consists of performing pansharpening of the PAN image with the low-resolution MSI data using standard pansharpening algorithm, in this case the Brovey method, to obtain high-resolution MSI images, which are then used to train \namemodel. This strategy corresponds to the original EOGS method.}
%   {shows different ways of handling different image spectra, comparing the \textit{Brovey} method, the \textit{linear combination} method and the direct rendering method. \\The Brovey method is 
% a pansharpening method used by EOGS as a data preprocessing step. It pansharpen the high-resolution PAN images with the low-resolution MSI images, obtaining high-resolution RGB images.}
 Table~\ref{tab:PAN_quantitiative} summarizes the results obtained with different strategies for handling panchromatic and MSI data, comparing the \textit{Brovey} pansharpening method, the \textit{linear combination} method, the \textit{single channel} and the \panstrategy method defined earlier.
 The Brovey method corresponds to the classical Brovey pansharpening technique, which is used in EOGS as a preprocessing step. It fuses the high-resolution PAN image with the low-resolution MSI image to produce a high-resolution MSI image, which is then used as the sole input for training \namemodel.

  The second approach, referred to as \textit{linear combination}, involves first rendering a high-resolution MSI image \( I^{\mathcal{A}_n}_{\text{MSI}} \), from which the corresponding PAN image \( I^{\mathcal{A}_n} \) is derived through an ``MSI-to-PAN'' linear combination defined as 
\( I^{\mathcal{A}_n} = \vec{\alpha}~\cdot~I^{\mathcal{A}_n}_{\text{MSI}} \),
where \( \vec{\alpha}~\in~\mathbb{R}^3 \) are fixed parameters pre-estimated using standard linear regression~\cite{Eonerfpansharp}. In this configuration, training is explicitly performed using both the PAN and low-resolution MSI inputs, by computing the loss between the synthesized and training PAN images, as well as between a rendered low-resolution MSI and its corresponding training data.

The third approach, single channel, omits the low-resolution MSI inputs as in the \panstrategy configuration. However, it reconstructs the scene directly from PAN images only, such that the color features $\vec{f}_k$ lie in $\R^d$ with $d=1$. 

The final approach, \panstrategy, uses only the panchromatic data and still achieves reconstruction accuracy comparable to the Brovey configuration. In contrast, the linear combination method, which jointly trains on both MSI and panchromatic images, leads to a noticeable performance degradation of approximately 0.5 m.
When comparing the single channel and 3-PAN approaches, both operating on panchromatic data only, the single-channel strategy produces slightly worse results (by approximately 0.14~m on average). This observation suggests that the modeling of panchromatic rendering has a non-negligible impact on reconstruction quality.
Although the applied pansharpening technique is the Brovey method, these findings indicate that Gaussian Splatting cannot effectively exploit both modalities simultaneously during training. Instead, panchromatic and MSI inputs should be processed separately. This observation is consistent with previous work~\cite{sprintson_fusionrf_2025}, where the two modalities were also treated independently.

\begin{figure}[!t]
\centering
\setlength{\tabcolsep}{3pt}
\def\arraystretch{0.0}
\begin{tabular}{@{}cc@{}}
{\footnotesize None} & {\footnotesize \namelearnwv} \\[1.2pt]
\includegraphics[width=0.34\linewidth]{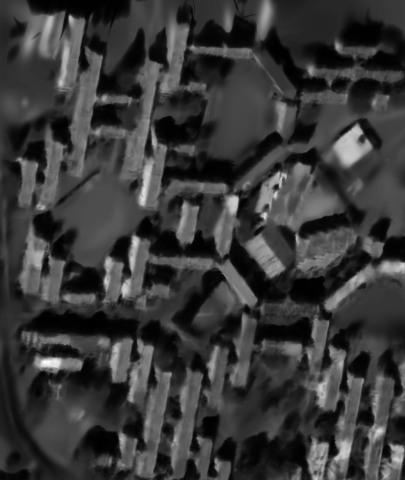} &
\includegraphics[width=0.34\linewidth]{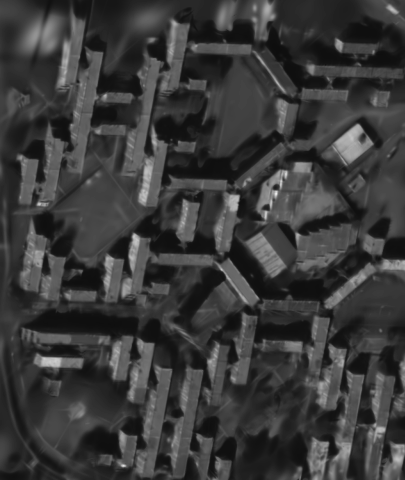}\\[2.4pt]
{\footnotesize Optical flow} & {\footnotesize External B.A.} \\[1.2pt]
\includegraphics[width=0.34\linewidth]{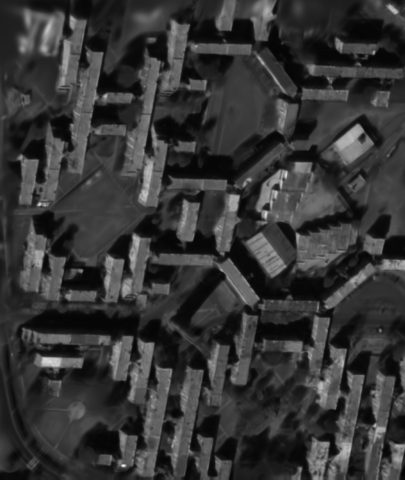} &
\includegraphics[width=0.34\linewidth]{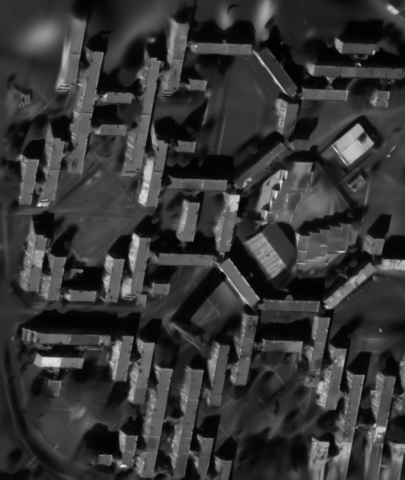}
\end{tabular}
\vspace{-.5em}
\caption{Rendered images on IARPA 001 using different B.A. strategies. Notice how the images become sharper.}
\vspace{0.5cm}
% \caption{From top-left to bottom-right: rendered images on IARPA\_001 using the raw RPC, the \namelearnwv method, the optical flow correction, and the bundle-adjusted cameras on the panchromatic images. Notice how the images become sharper.}
\label{fig:RPC_qualitative_comparison_rendering}
\end{figure}

% Thin version of Table 2
\begin{table}[h] 
\centering
\setlength{\tabcolsep}{2pt}
\renewcommand{\arraystretch}{1.1}
\begin{tabular}{@{\extracolsep{4pt}}rcccccc@{}}
& \scriptsize{\rot[90]{HR MSI}} & \scriptsize{\rot[90]{LR MSI}} & \scriptsize{\rot[90]{HR PAN}} & JAX $\downarrow$ & IARPA $\downarrow$ & Mean $\downarrow$ \\ \hline
Brovey       & \cmark & \xmark & \xmark & \textbf{1.19} & \textbf{1.46} & \textbf{1.33} \\
Linear combination & \xmark & \cmark & \cmark & 1.95     & 1.58     & 1.77      \\
Single channel     & \xmark & \xmark & \cmark & 1.36     & 1.58     & 1.47      \\
3-PAN    & \xmark & \xmark & \cmark & \textbf{1.19} & 1.47     & 1.33      \\ \bottomrule

\end{tabular}
\vspace{-.5em}
\caption{Mean absolute error (MAE) on elevation for different pansharpening strategies. HR denotes high-resolution, and LR denotes low-resolution data. Lower values indicate better performance. For a fair comparison, all methods are trained for exactly \num{5000} iterations. 
}
\vspace{0.2cm}
\label{tab:PAN_quantitiative}
\end{table}

% \begin{figure}[h]
%   \centering
%   \includegraphics[width=0.38\linewidth]{figures/paper/res-ba/final_rpcraw_image.png}
%   \includegraphics[width=0.38\linewidth]{figures/paper/res-ba/final_learnwv_image.png}\\
%   \includegraphics[width=0.38\linewidth]{figures/paper/res-ba/final_flowmatch_image.png}
%   \includegraphics[width=0.38\linewidth]{figures/paper/res-ba/final_rpcba_image.png}
%   \caption{From top-left to bottom-right: rendered images on IARPA\_001 using the raw RPC, the \namelearnwv method, the optical flow correction, and the bundle-adjusted cameras on the panchromatic images. Notice how the images become sharper.}
%   \label{fig:RPC_qualitative_comparison_rendering}
% \end{figure}

\begin{figure}[!hb]
\centering
% Titres des colonnes
% \setlength{\tabcolsep}{0pt} % remove horizontal space between columns
% \renewcommand{\arraystretch}{0} % remove vertical space between rows
\setlength{\tabcolsep}{0.6pt}
\def\arraystretch{0.0}
\begin{tabular}{@{}cccc@{}}
GT Lidar DSM & EOGS & \namemodel & \\[4pt]
\includegraphics[width=0.32\linewidth]{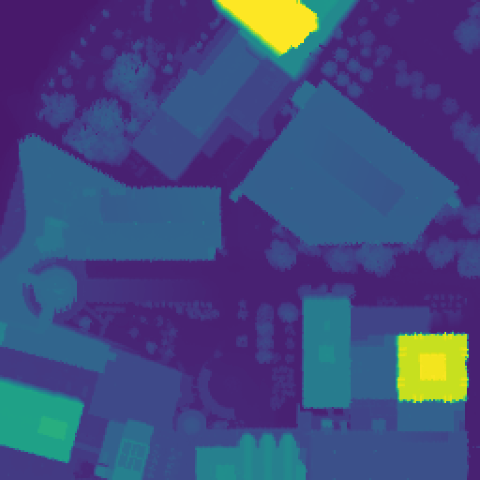}&
\includegraphics[width=0.32\linewidth]{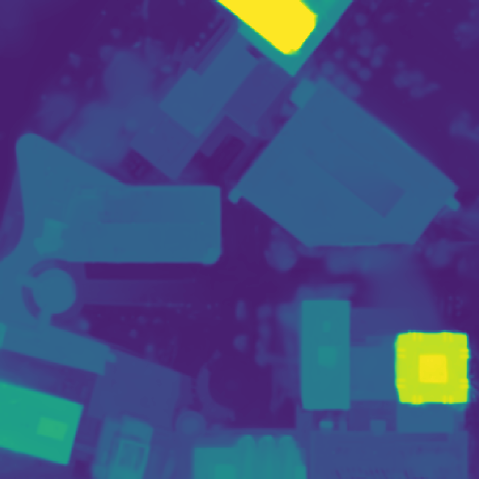}&
\includegraphics[width=0.32\linewidth]{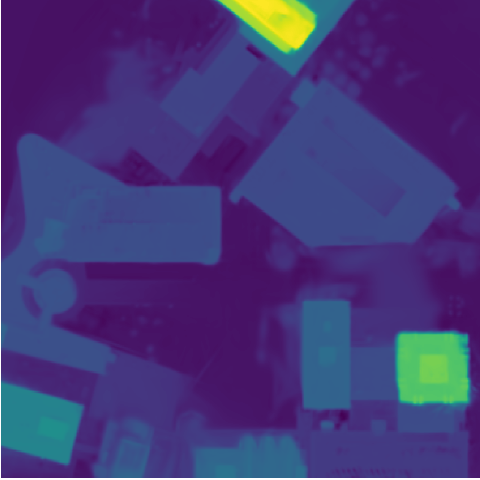}&
\includegraphics[width=0.15\linewidth]{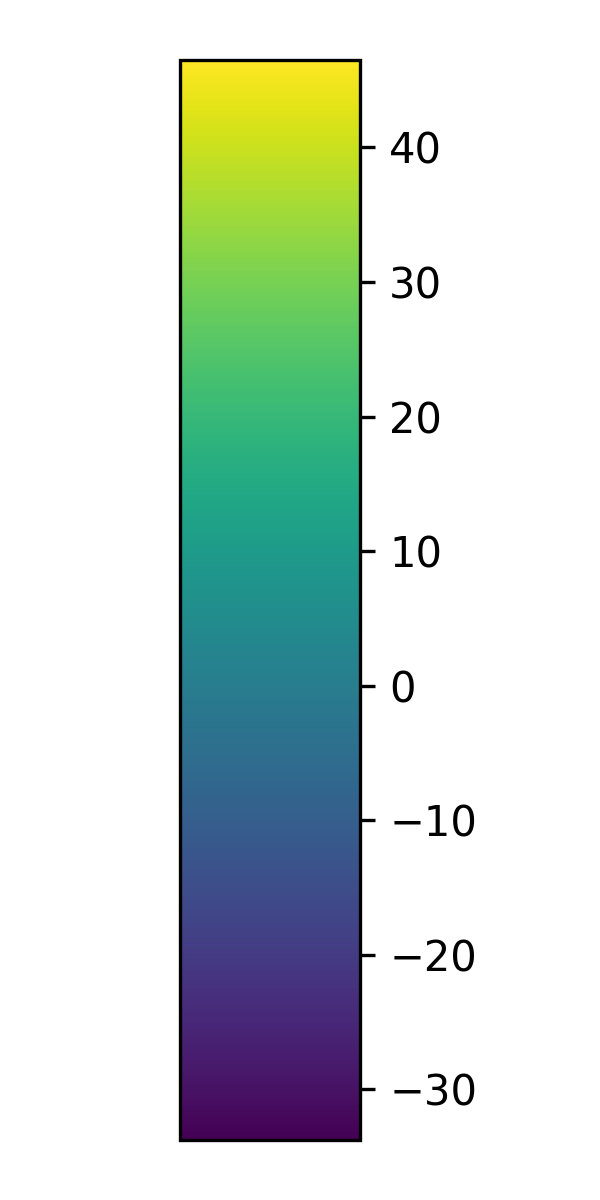} \\[1.2pt]

\includegraphics[width=0.32\linewidth]{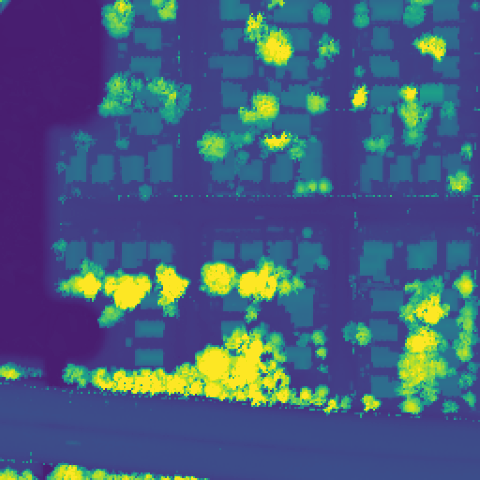}&
\includegraphics[width=0.32\linewidth]{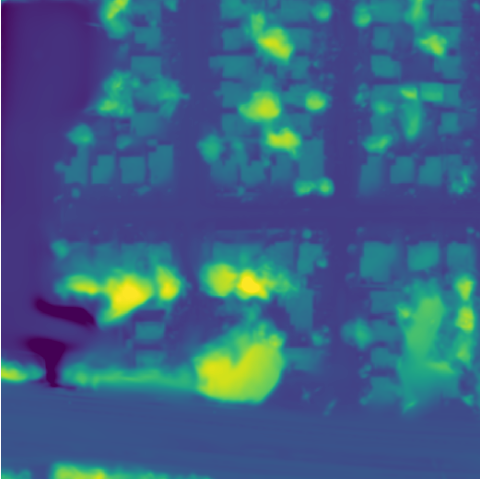}&
\includegraphics[width=0.32\linewidth]{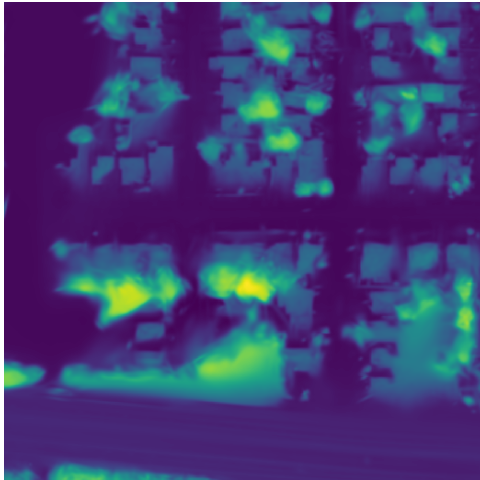}&
\includegraphics[width=0.15\linewidth]{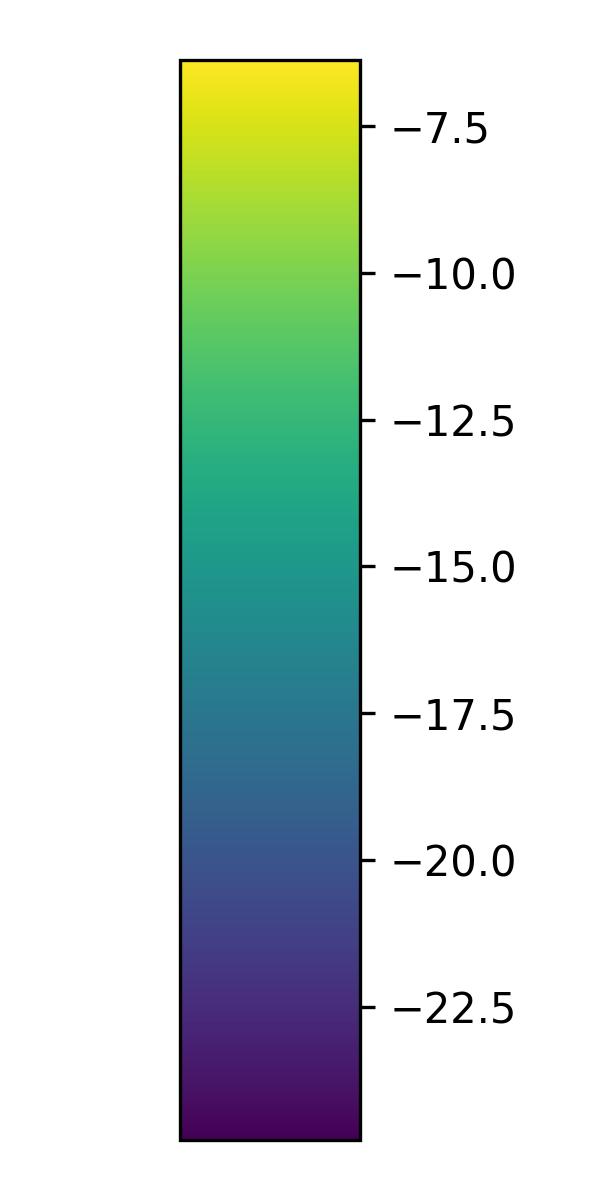} \\[1.2pt]

\includegraphics[width=0.32\linewidth]{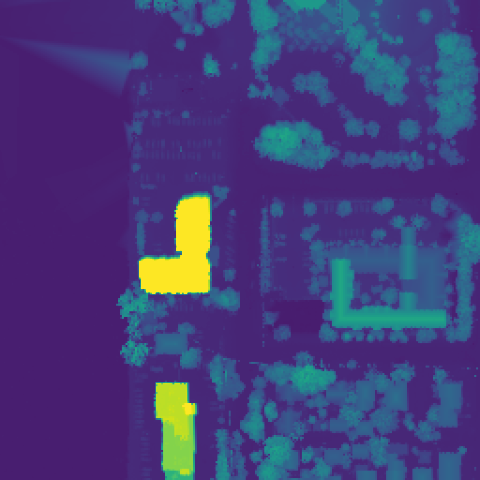}&
\includegraphics[width=0.32\linewidth]{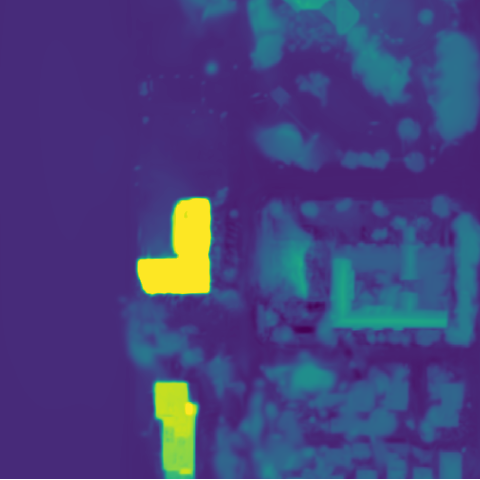}&
\includegraphics[width=0.32\linewidth]{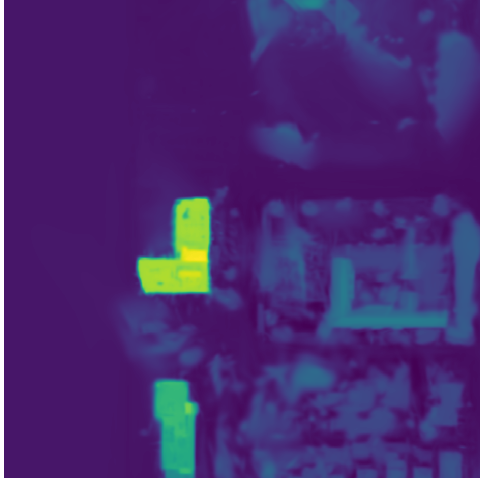}&
\includegraphics[width=0.15\linewidth]{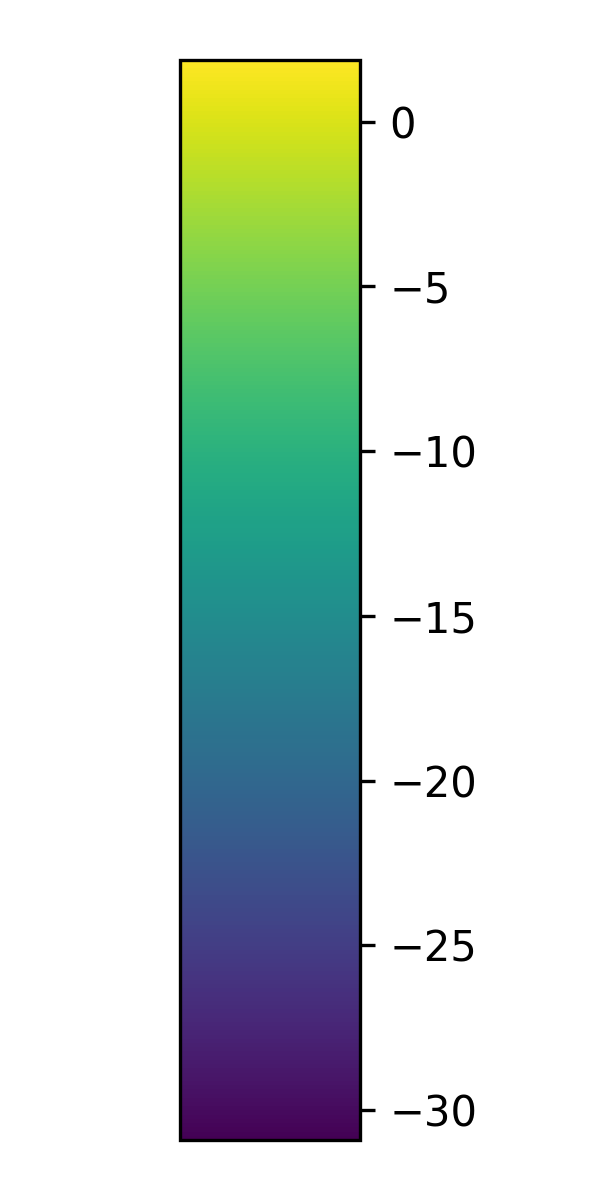}

\end{tabular}
\vspace{-.5em}
\caption{Qualitative comparison of EOGS and EOGS++ predicted DSMs on the JAX 214, JAX 004 and JAX 260 scenes.}
\label{fig:visual}
\end{figure}

\subsection{Comparing EOGS++ with the state of the art} % final method; Complete EOGS++ pipeline 

\begin{table*}[!h]
\centering
\setlength{\tabcolsep}{3pt}
\renewcommand{\arraystretch}{1.1}
\begin{tabular}{@{\extracolsep{4pt}}rccccccccccccc@{}}
\multirow{2}{*}{} 
& &
& \multicolumn{5}{c}{JAX} 
& \multicolumn{4}{c}{IARPA} 
& \multirow{2}{*}{\shortstack{Pooled\\mean $\downarrow$}} 
& \multirow{2}{*}{Time [min] $\downarrow$} \\ 
\cline{4-8} \cline{9-12}
% Some eldritch incantations for making table headers pretty :) 
& \adjustbox{angle=90,set height=0pt,}{\scriptsize \makecell[l]{B.A. RPC \\ Brovey Image}}
& \adjustbox{angle=90,set height=0pt,}{\scriptsize \makecell[l]{Opacity Reset \\ TSDF}}
& 004 & 068 & 214 & 260 & mean $\downarrow$ 
& 001 & 002 & 003 & mean $\downarrow$ 
& & \\ \hline
EOGS
& \cmark & \xmark 
& 0.77 & 1.07 & 1.65 & 1.26 & 1.19 
& 1.41 & 1.86 & 1.13 & 1.46 
& 1.33 & \textbf{9} \\
EOGS (40k iters)
& \cmark & \xmark 
& \textbf{0.70} & 1.03 & 1.65 & 1.27 & 1.16 
& 1.39 & 1.91 & \textbf{1.09} & 1.46 
& 1.31 & 72 \\
SAT-NGP 
& \cmark & \xmark 
& 1.03 & 1.26 & 2.17 & 1.43 & 1.47 
& 1.34 & 1.85 & 1.62 & 1.60 
& 1.53 & 25 \\
EO-NeRF 
& \cmark & \xmark 
& 1.02 & 1.03 & 1.55 & 1.24 & 1.21 
& 1.32 & 1.63 & 1.18 & 1.38 
& 1.29 & 900 \\
EOGS++ 
& \cmark & \cmark 
& 0.73 & 0.95 & \textbf{1.38} & \textbf{1.02} & \textbf{1.02} 
& \textbf{1.31} & \textbf{1.51} & 1.23 & \textbf{1.35} 
& \textbf{1.19} & 25 \\
EOGS++ 
& \xmark & \cmark 
& 0.72 & \textbf{0.90} & 1.64 & 1.16 & 1.10 
& 1.42 & 1.61 & 1.42 & 1.49 
& 1.29 & 32 \\ \midrule
Ablation %rpcraw of no fancy stuff repeat GT
& \xmark & \xmark & 0.88 & 1.25 & 1.63 & 1.46 & 1.30 & 1.57 & 1.93 & 1.33 & 1.61 & 1.46 & 31\\
\hline
\end{tabular}
\caption{Quantitative comparison of \namemodel and baseline methods on elevation reconstruction accuracy and runtime.Baselines include  EOGS~\protect\cite{savant2025EOGS}, SAT-NGP~\protect\cite{billouard2024satngp}, and EO-NeRF~\protect\cite{eonerf}.  The first column denote the use of preprocessed data. The two last row reports the performance when training directly with the raw RPCs provided by the satellite vendor and raw panchromatic data.}
\label{tab:main_result}
\end{table*}

 The results are reported in Table~\ref{tab:main_result} and visual results are reported in Figure~\ref{fig:visual}. Additional results obtained without the foliage mask, as well as further comparative figures, are provided in the supplementary material.
\namemodel, when using the same data modalities as prior work, consistently achieves the best performance on buildings compared to existing models by 0.1 meters. 

The early stopping mechanism effectively removes misplaced Gaussians and results in a more coherent scene reconstruction.

Instead, when relying solely on the non-bundle-adjusted panchromatic data, the model achieves performance comparable to the original EOGS, despite EOGS relying on camera priors and pre-pansharpened MSI data.

However, when evaluating results without the foliage mask, performance remains inferior to Eo-NeRF. As illustrated in Figure~\ref{fig:visual}, certain fine vegetation structures, particularly in the JAX 260 scene, are not fully preserved. As components such as early stopping and opacity reset act as regularization mechanisms, they tend to suppress high-frequency elements with limited geometric stability. Given that vegetation is inherently transient and varies with seasons and acquisition dates, these details are therefore less critical for our primary objective, which is the reconstruction of permanent structures such as buildings.

The last row of Table~\ref{tab:main_result} reports the performance of EOGS++ when trained directly on raw panchromatic data and raw RPCs, without any of the proposed enhancements. As PAN imagery provides weaker geometric cues than pansharpened inputs, the resulting MAE is higher than when using pansharpened data (shown  in table~\ref{tab:RPC_quantitative}). Nevertheless, integrating the full set of improvements yields a substantial reduction in error (about 0.2 m on average), indicating that each component contributes meaningfully to training stability and reconstruction accuracy.

}

\section{Conclusion}
In this work, we presented \namemodel, an enhanced Gaussian Splatting framework specifically designed for Earth observation. Building upon the original EOGS, our approach eliminates preprocessing requirements, such as external bundle adjustment and pansharpening, by directly integrating pose refinement through optical flow and by operating natively on raw panchromatic imagery. Additional mechanisms, including opacity reset, early stopping, and TSDF-based postprocessing, further improve reconstruction sharpness and geometric accuracy. 

Experiments conducted on the IARPA2016 and DFC2019 datasets demonstrate that EOGS++ achieves state-of-the-art reconstruction accuracy. The framework performs robustly across varying acquisition conditions, achieving consistent improvements over existing methods, though challenges remain in capturing fine vegetation structures.

\section*{Acknowledgments}
\vspace{-1em}

This work was granted access to the HPC resources of IDRIS under the allocation AD011012453R4 made by GENCI.

\vspace{-1em}

{
	\begin{spacing}{1.17}
		\normalsize
		\bibliography{ISPRSguidelines_authors} % Include your own bibliography (*.bib), style is given in isprs.cls
	\end{spacing}
}

\end{document}